\documentclass{article}
\usepackage{ijcai21}

\usepackage{times}
\usepackage{soul}
\usepackage{url}
\usepackage[hidelinks]{hyperref}
\usepackage[utf8]{inputenc}
\usepackage[small]{caption}
\usepackage{graphicx}
\usepackage{amsmath}
\usepackage{amsthm}
\usepackage{booktabs}
\usepackage{algorithm}
\usepackage{algorithmic}
\usepackage{enumitem}

\usepackage{natbib}

\newcommand{\politihop}{$\mathtt{PolitiHop}$}

\title{Multi-Hop Fact Checking of Political Claims}

\author{
Wojciech Ostrowski\footnote{Contact Author}\and
Arnav Arora\and
Pepa Atanasova\And
Isabelle Augenstein\\
\affiliations
Department of Computer Science, University of Copenhagen, Denmark\\
\emails
qnj566@alumni.ku.dk,
\{aar, pepa, augenstein\}@di.ku.dk
}

\begin{document}
\maketitle

\begin{abstract}
Recent work has proposed multi-hop models and datasets for studying complex natural language reasoning. One notable task requiring multi-hop reasoning is fact checking, where a set of connected evidence pieces leads to the final verdict of a claim. 
However, existing datasets either do not provide annotations for gold evidence pages, or the only dataset which does (FEVER) mostly consists of claims which can be fact-checked with simple reasoning and is constructed artificially.
Here, we study more complex claim verification of naturally occurring claims with multiple hops over interconnected evidence chunks. We: 1) construct a small annotated dataset, PolitiHop\footnote{We make the \politihop\ dataset and the code for the experiments publicly available on \href{https://github.com/copenlu/politihop}{https://github.com/copenlu/politihop} .}, of evidence sentences for claim verification; 2) compare it to existing multi-hop datasets; and 3) study how to transfer knowledge from more extensive in- and out-of-domain resources to PolitiHop. We find that the task is complex and achieve the best performance with an architecture that specifically models reasoning over evidence pieces in combination with in-domain transfer learning.  

\end{abstract}

\section{Introduction}
\noindent 

\begin{figure}[t]
    \centering
    \includegraphics[scale=0.65]{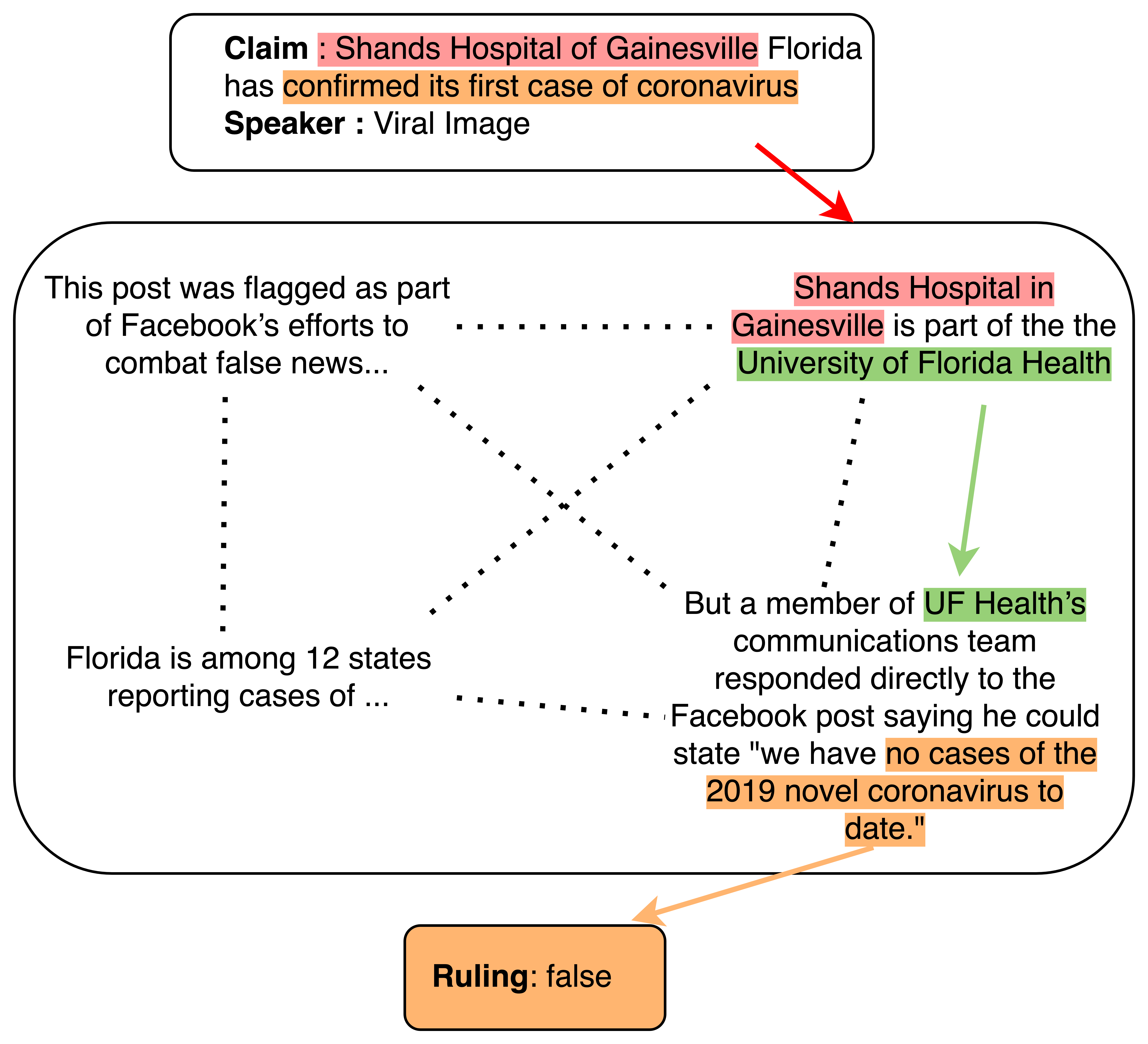}
    \caption{An illustration of multiple hops over an instance from \politihop. Each instance consists of a claim, a speaker
    , a veracity label, and a PolitiFact article the annotated evidence sentences. The highlighted sentences represent the evidence sentences a model needs to connect to arrive at the correct veracity prediction.}
    \label{fig:example}
\end{figure}

Recent progress in machine learning has seen interest in automating complex reasoning, where a conclusion can be reached only after following logically connected arguments.
To this end, multi-hop datasets and models have been introduced, which learn to combine information from several sentences to arrive at an answer. While most of them concentrate on question answering, fact checking is another task that often requires a combination of multiple evidence pieces to predict a claim's veracity. 

Existing fact checking models usually optimize only the veracity prediction objective and assume that the task requires a single inference step. Such models ignore that often several linked evidence chunks have to be explicitly retrieved and combined to make the correct veracity prediction. Moreover, they do not provide explanations of their decision-making, which is an essential part of fact checking.

\citet{fc-expl} note the importance of providing explanations for fact checking verdicts, and propose an extractive summarization model, which optimizes a ROUGE score metric w.r.t. a gold explanation. 
Gold explanations for this are obtained from the LIAR-PLUS~\citep{alhindi-etal-2018-evidence} dataset, which is constructed from PolitiFact\footnote{\url{https://www.politifact.com/}} articles written by professional fact checking journalists.
However, the dataset does not provide guidance on the several relevant evidence pieces that have to be linked and assume that the explanation requires a single reasoning step. FEVER~\citep{fever} is another fact checking dataset, which contains annotations of evidence sentences from Wikipedia pages. However, it consists of manually augmented claims, which require limited reasoning capabilities for verification as the evidence mostly consists of one or two sentences.

To provide guidance for the multi-hop reasoning process of a claim's verification and facilitate progress on explainable fact checking, we introduce \politihop, a dataset of 500 real-world claims with manual annotations of sets of interlinked evidence chunks from PolitiFact articles needed to predict the claims' labels. We provide insights from the annotation process, indicating that fact checking real-world claims is an elaborate process requiring multiple hops over evidence chunks, where multiple evidence sets are also possible.

To assess the difficulty of the task, we conduct experiments with lexical baselines, as well as a single-inference step model -- BERT~\citep{devlin2019bert}, and a multi-hop model -- Transformer-XH~\citep{trans-xh}. Transformer-XH allows for the sharing of information between sentences located anywhere in the document by eXtra Hop attention and achieves the best performance.
We further study whether multi-hop reasoning learned with Transformer-XH can be transferred to \politihop. We find that the model cannot leverage any reasoning skills from training on FEVER, while training on LIAR-PLUS improves the performance on \politihop. We hypothesize that this is partly due to a domain discrepancy, as FEVER is constructed from Wikipedia and consists of claims requiring only one or two hops for verification. In contrast, LIAR-PLUS is based on PolitiFact, same as \politihop.

Finally, we perform a detailed error analysis to understand the models' shortcomings and recognize possible areas for improvement. We find that the models perform worse when the gold evidence sets are larger and that, surprisingly, named entity (NE) overlap between evidence and non-evidence sentences does not have a negative effect on either evidence retrieval or label prediction. The best results for Transformer-XH on the dev and test sets are for a different number of hops -- 2 and 6, indicating that having a fixed parameter for the number of hops is a downside of Transformer-XH; this should instead be learned for each claim. Overall, our experiments constitute a solid basis to be used for future developments.




To summarise, our \textbf{contributions} are as follows:
\begin{itemize}[nosep]
\item We document the first study on multi-hop fact checking of political claims
\item We create a dataset for the task
\item We study whether reasoning skills learned with a multi-hop model on similar datasets can be transferred to \politihop
\item We analyze to what degree existing multi-hop reasoning methods are suitable for the task
\end{itemize}

\section{Multi-Hop Fact Checking}


A multi-hop fact checking model $f(X)$ receives as an input $X=\{(claim_i, document_i)| i\in[1,|X|]\}$, where $document_i = [sentence_{ij}|j \in [1, |document_i|]]$ is the corresponding PolitiFact article for $claim_i$ and consists of a list of sentences. During the training process, the model learns to (i) select which sentences from the input contain evidence needed for the veracity prediction $y_i^S = [y_{ij}^S \in \{0, 1\}|j \in [1, |document_i|]]$ (\textbf{sentence selection task}), where 1 indicates that the sentence is selected as an evidence sentence; and (ii) predict the veracity label of the claim $y_i^L \in \{True, False, Half-True\}$, based on the extracted evidence (\textbf{veracity prediction task}). The sentences selected by the model as evidence provide \textit{sufficient} explanation, which allows to verify the corresponding claim \textit{efficiently} instead of reading the whole article. Each evidence set consists of $k$ sentences, where $k \in [1, max_{i\in[1,|X|]}(|document_i|)]$ is a hyper-parameter of the model. Figure~\ref{fig:example} illustrates the process of multi-hop fact checking, where multiple evidence sentences provide different information, which needs to be connected in logical order to reach the final veracity verdict.

\begin{table}
\footnotesize
\centering
\begin{tabular}{p{3.7cm}rr}
\toprule
\bf Statistic & \bf Test & \bf Train\\
\midrule
\#Words per article & 569 (280.8) & 573 (269.1)\\
\#Sent. per article & 28 (12.8) & 28 (12.8)\\
\#Evidence sent. per article & 11.75 (5.56) & 6.33 (2.98)\\
\#Evidence sent. per set & 2.88 (1.43) & 2.59 (1.51)\\
\#Sets per article & 4.08 (1.83) & 2.44 (1.28)\\
\midrule
\multicolumn{3}{c}{\textbf{Label Distribution}}\\
False & 149 & 216\\
Half-true & 30 & 47\\
True & 21 & 37\\
\bottomrule
\end{tabular}

\caption{\politihop\ dataset statistics. Test set statistics are calculated for a union of two annotators; train instances are annotated by one annotator only, which makes some measures different across splits. We report the mean and standard deviation (in parentheses).} 
\label{table:data}
\end{table}


\subsection{Dataset}
We present \politihop, the first dataset for multi-hop fact checking of real-world claims. It consists of 500 manually annotated claims in written English, split into a training (300 instances) and a test set (200 instances). For each claim, the corresponding PolitiFact article was retrieved, which consists of a discussion of each claim and its veracity, written by a professional fact checker. The annotators then selected sufficient sets of evidence sentences from said articles.
As sometimes more than one set can be found to describe a reason behind the veracity of a claim independently, we further take each set in the training split as a separate instance, resulting in 733 training examples.
Each training example is annotated by one annotator, whereas each test example is annotated by two. We split the training data into train and dev datasets, where the former has 592 examples and the latter -- 141. For veracity prediction, we arrived at Krippendorf's $\alpha$ and Fleiss' $\kappa$ agreement values of 0.638 and 0.637, respectively. By comparison, \citet{fever} reported Fleiss' $\kappa$ of 0.684 on FEVER. For the sentence prediction, we attain Krippendorf's $\alpha$ of 0.437. A more in-depth description of the annotation process can be found in the appendix. 

Table~\ref{table:data} presents statistics of the dataset. 
The average number of evidence sentences per set is above 2, which already indicates that the task is more complex than the FEVER dataset. In FEVER, 83.2\% of the claims require one sentence, whereas in \politihop, only 24.8\% require one sentence. 




\section{Models}
We compare the performance of five different models to measure the difficulty of automating the task. 

\paragraph{Majority.}
Label prediction only. The majority baseline labels all claims as false.

\paragraph{Random.}
We pick a random number $k \in [1, 10]$ and then randomly choose $k$ sentences from the document as evidence. For label prediction, we randomly pick one of the labels. 

\paragraph{TF-IDF.}
For each instance $x_i$ we construct a vector $v_i^C=[v_{il}^C| l \in [0, |N^C|]]$ with TF-IDF scores $v_{il}^C$ for all n-grams $N^C$ found in all of the claims; and one vector $v_i^D=[v_{im}^D| m \in [0, |N^D|]]$ with TF-IDF scores $v_{im}^D$ for all n-grams $N^D$ found in all of the documents, where $n \in [2, 3]$. We then train a Naive Bayes model $g(V)$, where $V = \{v_i = (v_i^C \cdot v_i^D) | i \in [0, |X|]\}$ is the concatenation of the two feature vectors. We also remove English stop words using the built-in list in the Scikit-learn library~\citep{scikit-learn}.

\paragraph{BERT.}
We first train a Transformer model~\citep{vaswani2017attention}, which does not include a multi-hop mechanism, but applies a single inference step to both the evidence retrieval and the label prediction tasks. We employ BERT~\citep{devlin2019bert} with the base pre-trained weights. Each sentence from a fact checking document is encoded separately, combined with the claim and the author of the claim. We refer to the encoded triple as \textit{node $\tau$}. The tokens of one node $x_{\tau} = \{x_{\tau, j} | j \in [0, |x_{\tau}|]\}$ are encoded with the BERT model into contextualized distributed representations: $h_{\tau} = \{h_{\tau,j} | j \in [0, |x_{\tau}|\}$. The encoded representations of all nodes are passed through two feed-forward layers:
\begin{gather}
\small
p(y^{L}|\tau) = softmax(Linear(h_{\tau,0})) \label{eq:lone}\\
p(y^{S}|\tau) = softmax(Linear(h_{\tau,0})) \label{eq:ltwo}\\
p(y^{L}|X) = \sum_{\tau}p(y^{L}|\tau)p(y^{S}|\tau) \label{eq:final}
\end{gather}
The first layer predicts the veracity of the claim given a particular node $\tau$ by using the contextual representation of the ``[CLS]'' token, located at the first position (Eq.~\ref{eq:lone}). The second feed-forward layer learns the importance of each node in the graph (Eq.~\ref{eq:ltwo}). The outputs of these two layers are combined for the final label prediction (Eq.~\ref{eq:final}).
For evidence prediction, we choose $k$ most important sentences, as ranked by the second linear layer. In our experiments, we set $k=6$ since this is the average number of evidence sentences selected by a single annotator.
The implementation of the feed-forward prediction layers is the same as in Transformer-XH, described below, and can be viewed as an ablation of Transformer-XH removing the eXtra Hop attention layers.

\paragraph{Transformer-XH.}
Transformer-XH is a good candidate for a multi-hop model for our setup as it has previously achieved the best multi-hop evidence retrieval results on FEVER. It is also inspired by and improves over other multi-hop architectures~\citep{kgat,zhou-etal-2019-gear}, and we conjecture that the results should be generalisable for its predecessors as well. Not least, its architecture allows for ablation studies of the multi-hop mechanism. 
Following previous work on applying Transformer-XH to FEVER \citep{trans-xh}, we encode node representations as with the BERT model and construct a fully connected graph with them. Transformer-XH uses eXtra hop attention layers to enable information sharing between the nodes.
An eXtra hop attention layer is a Graph Attention layer (GAT) \citep{gat}, which receives as input a graph $\{X, E\}$ of all evidence nodes $X$ and the edges between them $E$, where the edges encode the attention between two nodes in the graph. Each eXtra hop layer computes the attention between a node and its neighbors, which corresponds to one hop of reasoning across nodes.
Transformer-XH applies L eXtra hop layers to the BERT node encodings $H_0$, which results in new representations $H_L$ that encode the information shared between the nodes, unlike BERT, which encodes each input sentence separately.
We use three eXtra hop layers as in \citep{trans-xh}, which corresponds to three-hop reasoning, and we experiment with varying the number of hops.
The representations $H_L$ are passed to the final two linear layers for label and evidence prediction as in BERT. The final prediction of the veracity label $p(y^{L}|\{X, E\})$ now can also leverage information exchanged in multiple hops between the nodes through the edges $E$ between them.

\section{Experiments}
We address the following research questions:
\begin{itemize}[nosep]
\item Can multi-hop architectures successfully reason over evidence sets on \politihop? 
\item How do multi-hop vs. single inference architectures fare in an adversarial evaluation, where named entities (NE) in evidence and non-evidence sentences overlap? 
\item Does pre-training on related small in-domain or large out-of-domain datasets improve model performance? 
\end{itemize}

We further perform ablation studies to investigate the influence of different factors on performance (see Section~\ref{sec:discussion}).

\subsection{Experimental Setup}
\paragraph{Metrics.} We use macro-F1 score and accuracy for the veracity prediction task and F1 and precision for the evidence retrieval task. To calculate the performance on both tasks jointly, we use the FEVER score \citep{fever}, where the model has to retrieve at least one full evidence set and predict the veracity label correctly for the label prediction to count as correct. We consider a single evidence set to be sufficient for correct label prediction. As each example from train and dev sets in \politihop, and every example from LIAR-PLUS, has one evidence set, all evidence sentences need to be retrieved for these. The employed measures for evidence retrieval allow for comparison to related work and for relaxing the requirements on the models. We consider the FEVER score to be the best for evaluating explainable fact checking.

\paragraph{Dataset Settings.}
We consider three settings: $\mathtt{full}$ article, $\mathtt{even}$ $\mathtt{split}$ and $\mathtt{adversarial}$.
For $\mathtt{full}$, the whole article for each claim is given as input. 
For $\mathtt{even}$ $\mathtt{split}$, we pick all sentences from the same article, but restrict the number of non-evidence sentences to be at most equal to the number of evidence sentences. Non-evidence sentences are picked randomly. 
This results in a roughly even split between evidence and non-evidence sentences for the test set. 
Since we divide train and dev datasets into one evidence set per example, but keep all non-evidence for each, the number of non-evidence sentences for instances in these splits is usually 2-3 times larger than the number of evidence sentences. To examine if the investigated multi-hop models overfit on named entity (NE) overlaps, we further construct an $\mathtt{adversarial}$ dataset from the even split dataset by changing each non-evidence sentence to a random sentence from any PolitiFact article, which contains at least one NE present in the original evidence sentences. While such sentences can share information about a relevant NE, they are irrelevant for the claim. We argue that this is a good testbed to understand if a fact checking model can successfully reason over evidence sets and identify non-evidence sentences, even if they contain relevant NEs, which are rather surface features not indicating whether the sentence is relevant to the claim.

\paragraph{Training Settings.} We perform transfer learning, training on in-domain data (LIAR-PLUS, \politihop), out-of-domain data (FEVER), or a combination thereof. See the appendix for details on training regimes and hyper-parameters.

Note that the measures do not consider the order of the sentences in the evidence set, and the systems do not predict that as well. We believe other measures and models that take that into account should be explored in future work. Here we consider them to appear in the same order as in the document. This also corresponds to the way they were annotated.

\begin{table*}[t]
\small
\centering
    \begin{tabular}{lrr|rr|r|rr|rr|rr|rr|r}
      \toprule
      & \multicolumn{5}{c|}{\textbf{Dev}} & \multicolumn{5}{c}{\textbf{Test}}\\
      & \bf L-F1 & \bf L-Acc & \bf E-F1 & \bf E-Prec & \bf FEVER &
      \bf 
      L-F1 & \bf L-Acc & \bf E-F1 & \bf E-Prec & \bf FEVER\\
      \midrule
       Random & 34.1 & 38.5 & 22.9 & 30.2 & 4.5
      & 24.2 & 27.7 & 14.7 & 12.2 & 0.7\\
      Majority & 27.3 & 69.5 & - & - & - & 28.6 & 75.0 & - & - & -\\
      Annotator & - & - & - & - & - & 76.3 & - & 52.4 & 49.2 & -\\
      TF-IDF & 34.4 & 69.5 & - & - & - & 34.0 & 76.0 & - & - & -\\
      \midrule
      \multicolumn{11}{c}{\textbf{LIAR-PLUS full articles dataset}} \\
      BERT & 45.4 & 70.9 & \bf 18.4 & \bf 13.7 & \bf 14.9
      & \bf 57.0 & 76.0 & \bf 32.9 & \bf 38.9 & \bf 13.0\\
      Transformer-XH & \bf 56.2 & \bf 74.5 & 17.1 & 12.8 & 14.2
      & 56.3 & \bf 79.5 & 30.3 & 35.8 & 12.0\\
     \midrule
     \multicolumn{11}{c}{\textbf{PolitiHop full articles dataset}} \\
     BERT & 54.7 & 69.5 & \bf 32.0 & \bf 23.6 & 31.9
      & \bf 44.8 & \bf 76.0 & \bf 47.0 & \bf 54.2 & \bf \underline{24.5}\\
      Transformer-XH & \bf 61.1 & \bf 76.6 & 30.4 & 22.3 & \bf 34.8
      & 43.3 & 75.5 & 44.7 & 51.7 & 23.5\\
      \midrule
    \multicolumn{11}{c}{\textbf{LIAR-PLUS and PolitiHop full articles}} \\
      BERT & 64.4 & 75.9 & 29.6 & 21.7 & 28.4
      & \bf \underline{57.8} & 79.5 & 45.1 & 52.2 & 23.5\\
      Transformer-XH & \bf \underline{64.6} & \bf \underline{78.7} & \bf \underline{32.4} & \bf \underline{23.8} & \bf \underline{38.3}
      & 57.3 & \bf \underline{80.5} & \bf \underline{47.2} & \bf \underline{54.5} & \bf \underline{24.5}\\
      \bottomrule
    \end{tabular}
  
  \caption{\politihop\ results for label (L), evidence (E) and joint (FEVER) performance in the $\mathtt{full}$ setting. Best results with a particular training dataset (LIAR-PLUS/PolitiHop/LIAR-PLUS and PolitiHop) are emboldened and the best results across all set-ups are underlined.} 
  \label{table:baseline}
\end{table*}

\section{Results}
\label{sec:results}

\paragraph{Full article setting.} From the results
in Table~\ref{table:baseline}, we can observe that both BERT and Transformer-XH greatly outperform the Random and TF-IDF baselines. 
Out of BERT and Transformer-XH, neither model clearly outperforms the other on our dataset. This is surprising as Transformer-XH outperforms the BERT baselines by a significant margin on both FEVER and the multi-hop dataset HotpotQA \citep{trans-xh}. 
However, we observe that the best performance is achieved with Transformer-XH trained on LIAR-PLUS, then fine-tuned on \politihop. It also achieves the highest FEVER scores on \politihop\ in that setting. 
Further, very low FEVER scores of both Transformer-XH and BERT indicate how challenging it is to retrieve the whole evidence set.

\paragraph{Adversarial setting.} 
We train the Transformer-XH models on the $\mathtt{even}$ $\mathtt{split}$ setting, then evaluate it on both  $\mathtt{adversarial}$ and $\mathtt{even}$ $\mathtt{split}$ datasets (see Table~\ref{table:adv}). 
The model performs similarly in both settings. When compared on test sets, it achieves a higher FEVER score on the $\mathtt{adversarial}$, but dev sets FEVER score is higher on the $\mathtt{even}$ $\mathtt{split}$ setting. Overall, the results show Transformer-XH is robust towards NE overlap.

\begin{table*}[t]
\small
\centering
    \begin{tabular}{lrr|rr|r|rr|rr|rr|rr|r}
      \toprule
      & \multicolumn{5}{c|}{\textbf{Dev}} & \multicolumn{5}{c}{\textbf{Test}}\\
      & \bf L-F1 & \bf L-Acc & \bf E-F1 & \bf E-Prec & \bf FEVER &
      \bf L-F1 & \bf L-Acc & \bf E-F1 & \bf E-Prec & \bf FEVER\\
      \midrule
      even split & \bf 58.1 & \bf 71.6 & 47.7 & 35.5 & \bf 52.5
      & \bf 62.9 & \bf 82.0 & 58.2 & 66.7 & 31.0\\
      adversarial & 56.5 & 70.9 & \bf 49.9 & \bf 38.7 & 46.8
      & 56.4 & 77.0 & \bf 63.6 & \bf 76.0 & \bf 33.5\\
      \bottomrule
    \end{tabular}
  
  \caption{\politihop\ $\mathtt{adversarial}$ vs $\mathtt{even}$ $\mathtt{split}$ dataset results for label (L), evidence (E) and joint (FEVER) performance for Transformer-XH trained on LIAR-PLUS and \politihop\ on the $\mathtt{even}$ $\mathtt{split}$ setting. Best result emboldened.}
  \label{table:adv}
\end{table*}

\paragraph{Out-of-domain pre-training on FEVER.}
In this experiment, we examine whether pre-training Transformer-XH on the large, but out-of-domain dataset FEVER, followed by fine-tuning on LIAR-PLUS, then on \politihop\, improves results on \politihop. 
As can be seen from Table~\ref{table:fever-pre}, it does not have a positive effect on performance in the $\mathtt{full}$ setting, unlike pre-training on LIAR-PLUS. We hypothesize that the benefits of using a larger dataset are outweighed by the downsides of it being out-of-domain.
We further quantify the domain differences between datasets. We use Jensen-Shannon divergence \citep{jlin}, commonly employed for this purpose \citep{conf/emnlp/RuderP17}. The divergence between FEVER and \politihop\ is 0.278, while between LIAR-PLUS and \politihop\ is 0.063, which further corroborates our hypothesis.
Another reason might be that \politihop\ has several times more input sentences compared to FEVER.
Labelling difference might matter as well: FEVER uses `true', `false' and `not enough info', while \politihop\ uses `true', `false' and `half-true'.

\begin{table*}[t]
\footnotesize
\centering
    \begin{tabular}{lrr|rr|r|rr|rr|rr|rr|r}
      \toprule
      & \multicolumn{5}{c|}{\textbf{Dev}} & \multicolumn{5}{c}{\textbf{Test}}\\
      & \bf L-F1 & \bf L-Acc & \bf E-F1 & \bf E-Prec & \bf FEVER &
      \bf L-F1 & \bf L-Acc & \bf E-F1 & \bf E-Prec & \bf FEVER\\
      \midrule
      FEVER+LIAR-PLUS+\politihop& 48.6 & 70.2 & 30.5 & 22.2 & 32.6
      & \bf 59.9 & \bf 83.0 & 45.1 & 52.7 & 21.5\\
      LIAR-PLUS+\politihop& \bf 64.6 & \bf 78.7 & \bf 32.4 & \bf 23.8 & \bf 38.3
      & 57.3 & 80.5 & \bf 47.2 & \bf 54.5 & \bf 24.5\\
      \bottomrule
    \end{tabular}
  
  \caption{\politihop\ $\mathtt{full}$ results for label (L), evidence (E) and joint (FEVER) performance for Transformer-XH trained on different datasets. Best model emboldened.} 
  \label{table:fever-pre}
\end{table*}

\section{Analysis and Discussion}
\label{sec:discussion}
In Section~\ref{sec:results}, we documented experimental results on multi-hop fact checking of political claims. Overall, we found that multi-hop training on Transformer-XH gives small improvements over BERT, that pre-training on in-domain data helps, and that Transformer-XH deals well with an adversarial test setting.
Below, we aim to further understand the impact of modeling multi-hop reasoning explicitly with a number of ablation studies:
\begin{itemize}[nosep]
\item How the evidence set size affects performance
\item Varying the hops' number in Transformer-XH
\item The impact of evidence set size on performance
\item How NE overlap affects performance
\item To what extent Transformer-XH pays attention to relevant evidence sentences
\end{itemize}

Further ablation studies can be found in the appendix, namely on: impact of varying the number of evidence sentences on evidence retrieval; how to weigh the different loss functions (for label vs. evidence prediction); if providing supervision for evidence sentence positions impacts performance; and to what degree high label confidence is an indication of high performance.

\paragraph{Varying the number of hops in Transformer-XH.}
We train Transformer-XH with a varying number of hops to see if there is any pattern in how many hops result in the best performance. \citet{trans-xh} perform a similar experiment and find that 3 hops are best, similar for 2-5 hops, while the decrease in performance is noticeable for 1 and 6 hops. 
We experiment with hops between 1 and 7 (see Table~\ref{table:hops}). Evidence retrieval performance is quite similar in each case. There are some differences for the label prediction task: 1 and 2 hops have slightly worse performance, the 4-hop model has the highest test score and the lowest dev score, while the exact opposite holds for the 5-hop model. Therefore, no clear pattern can be found. One reason for this could be the high variance of the annotated evidence sentences in \politihop. 

\begin{table*}[!ht]
\small
\centering
    \begin{tabular}{lrr|rr|r|rr|rr|rr|rr|r}
      \toprule
      & \multicolumn{5}{c|}{\textbf{Dev}} & \multicolumn{5}{c}{\textbf{Test}}\\
      & \bf L-F1 & \bf L-Acc & \bf E-F1 & \bf E-Prec & \bf FEVER &
      \bf L-F1 & \bf L-Acc & \bf E-F1 & \bf E-Prec & \bf FEVER\\
      \midrule
      1 & 54.0 & \bf 75.2 & 47.1 & 35.2 & 52.5
      & 58.9 & 79.5 & 58.7 & \bf 67.6 & 33.0\\
      2 & 56.1 & 73.0 & 47.5 & 35.4 & \bf 53.9
      & 59.8 & 78.5 & 58.1 & 66.7 & 32.5\\
      3 & 58.1 & 71.6 & \bf 47.7 & \bf 35.5 & 52.5
      & 62.9 & 82.0 & 58.2 & 66.7 & 31.0\\
      4 & 53.3 & 70.9 & 47.0 & 34.9 & 50.4
      & \bf 65.0 & \bf 82.0 & \bf 58.9 & \bf 67.6 & 33.0\\
      5 & \bf 59.6 & 73.0 & \bf 47.7 & \bf 35.5 & 51.8
      & 55.3 & 76.5 & 58.7 & 67.3 & 32.0\\
      6 & 56.5 & 73.0 & 45.9 & 34.2 & 50.4
      & 64.9 & 81.5 & 57.5 & 66.0 & \bf 35.0\\
      7 & 56.3 & 71.6 & 46.4 & 34.6 & 50.4
      & 62.8 & 81.5 & 57.9 & 66.4 & 33.0\\
      \bottomrule
    \end{tabular}
  \caption{\politihop\ Transformer-XH results for label (L), evidence (E) and joint (FEVER) performance for training on the LIAR-PLUS + \politihop\ $\mathtt{even}$ $\mathtt{split}$ datasets with a varying number of hop layers. Best sentence number emboldened.}
  \label{table:hops}
\end{table*}

\paragraph{Evidence set size vs. performance.}
Not surprisingly, larger number of evidence sentences leads to higher precision and lower recall, resulting in a lower FEVER score. This is true for both models, as Table~\ref{table:chain-len} (top) indicates. We also notice that the smaller the number, the smaller the ratio of evidence to non-evidence sentences.
For instance, if a claim has two sets of evidence, one of size 1 and the other of size 3, then after splitting into one example per set, there are 4 non-evidence sentences in each of the two examples, but the one with set of size 1 has only one evidence sentence -- which decreases the evidence to non-evidence ratio and makes it more difficult to achieve high precision.

\begin{table*}[!ht]
\small
\centering
    \begin{tabular}{lrr|rr|r|rr|rr|rr|rr|r}
      \toprule
      & \multicolumn{5}{c|}{\textbf{Transformer-XH}} & \multicolumn{5}{c}{\textbf{BERT}}\\
      & \textbf{L-F1} & \textbf{L-Acc} & \textbf{E-F1} & \textbf{E-Prec} & \textbf{FEVER} &
      \textbf{L-F1} & \textbf{L-Acc} & \textbf{E-F1} & \textbf{E-Prec} & \textbf{FEVER}\\
      \midrule
      1 or 2 evidence sentences & \bf 63.9 & \bf 76.8 & \bf 43.7 & \bf 29.2 & \bf 74.4
      & 53.5 & 72.0 & 41.3 & 27.5 & 62.2\\
      3+ evidence sentences & \bf 60.9 & 66.1 & \bf 67.5 & \bf 56.2 & \bf 42.4
      & 57.8 & 66.1 & 65.8 & 54.8 & 40.7\\
      \midrule
      $<$  40\% NE overlap & \bf 62.5 & \bf 77.0 & \bf 59.1 & \bf 46.2 & \bf 62.3
      & 62.0 & 75.4 & 57.7 & 45.1 & 59.0\\
      
      $\geq$ 40\% NE overlap & \bf 63.6 & \bf 71.0 & \bf 48.5 & \bf 35.5 & \bf 60.9
      & 47.5 & 66.7 & 46.2 & 33.8 & 49.3\\
      \bottomrule
    \end{tabular}
  \caption{\politihop\ $\mathtt{adversarial}$ dev set performance vs. (top) evidence set size and (bottom) NE overlap between evidence and non-evidence sentences for label (L), evidence (E) and joint (FEVER) performance. Better model emboldened.}
  \label{table:chain-len}
\end{table*}

\paragraph{Named entity overlap vs. performance.}
To measure the effect of having the same NEs in evidence and non-evidence sentences, we computed NE overlap -- a measure of the degree to which evidence and non evidence sentences share NEs. We compute the overlap as $|E \cap N|/|E \cup N|$; E and N are sets of NEs in evidence and non-evidence sentences, respectively.
Table~\ref{table:chain-len} (bottom) shows that a higher NE overlap results in more confusion when retrieving evidence sentences, but it does not have a significant influence on label prediction in the case of Transformer-XH. For BERT, higher NE overlap leads to a bigger, negative effect on both tasks. This suggests Transformer-XH is more robust to NE overlaps.

      

\begin{table}[ht]
\footnotesize
 \centering
 \setlength{\tabcolsep}{4pt}
    \begin{tabular}{cccc}
      \toprule
      \textbf{ev} $\rightarrow$ \textbf{non-ev} & \textbf{ev} $\rightarrow$ \textbf{ev} & \textbf{non-ev} $\rightarrow$ \textbf{non-ev} & \textbf{non-ev} $\rightarrow$ \textbf{ev}\\
      \midrule
      \bf 1.085 & 1.076 & 0.966 & 0.964\\
      \bottomrule
    \end{tabular}
  \caption{Attention weights in the last eXtra hop layer of Transformer-XH. The numbers are the average ratios of the actual attention weights to average attention weight of the given graph.}
  \label{table:attention}
\end{table}

\paragraph{Attention over evidence sentences.}
We investigate what attention patterns Transformer-XH learns. 
Ideally, attention flowing from evidence sentences should be higher than from non-evidence ones since this determines how much they contribute to the final representations of each sentence. To do this, we inspect the weights in the final eXtra hop layer. We normalize results by measuring the ratio of the given attention to the average attention for the given graph: 1 means average, over/under 1 means more/less than average.
Table~\ref{table:attention} shows average ratios for evidence vs. non-evidence sentences. One notable finding is that attention weights from evidence sentences are higher than average, and attention from non-evidence sentences is lower. The Welch t-test indicates that the difference is significant with a $p$-value lower than 10$^{-30}$. So, attention weights get more importance on average, but the magnitude of this effect is quite limited. This shows the limitations of using Transformer-XH for this task.



\section{Related Work}



\paragraph{Fact Checking.}
Several datasets have been released 
to assist in automating fact checking. 
\citet{vlachos-riedel-2014-fact} present a dataset with 106 political claim-verdict pairs. 
\citet{FNC} construct FakeNewsChallenge, a 50K headline-article pairs and formulate the task of fact checking as stance detection between the headline and the body of the article. The relationship between these two tasks is further explored in \citet{hardalov2021survey}.
\citet{Wang2017LiarLP} extract 12.8K claims from PolitiFact constituting the LIAR dataset.
\citet{alhindi-etal-2018-evidence} introduce the LIAR-PLUS dataset extending the latter with automatically extracted summaries from PolitiFact articles. These are, however, high-level explanations that omit evidence details. LIAR-PLUS also does not provide annotation of particular evidence sentences from the article leading to the final verdict and the possible different evidence sets. 
\citet{augenstein-etal-2019-multifc} present a real-world dataset constructed from 26 fact checking portals, including PolitiFact, consisting of 35k claims paired with crawled evidence documents.
\citet{fever} present the FEVER dataset, consisting of 185K claims produced by manually re-writing Wikipedia sentences. 
Furthermore, \citet{niewinski-etal-2019-gem} from the FEVER'2019 shared task~\citep{thorne-etal-2019-fever2} and \citet{hidey-etal-2020-deseption} use adversarial attacks to show the vulnerability of models trained on the FEVER dataset to claims that require more than one inference step.
Unlike prior work, we construct a dataset with annotations of the different reasoning sets and the multiple hops that constitute them.

\paragraph{Multi-Hop Datasets.}
Multi-hop reasoning has been mostly studied in the context of Question Answering (QA). 
\citet{hotpot} introduce HotpotQA with Wikipedia-based question-answer pairs requiring reasoning over multiple documents and provide gold labels for sentences supporting the answer. 
\citet{MedHop} introduce MedHop and WikiHop datasets for reasoning over multiple documents. These are constructed using Wikipedia and DrugBank as Knowledge Bases (KB), and are limited to entities and relations existing in the KB. This, in turn, limits the type of questions that can be generated. TriviaQA \citep{JoshiTriviaQA2017} and SearchQA \citep{searchqa} contain multiple documents for question-answer pairs but have few examples where reasoning over multiple paragraphs from different documents is necessary.

\paragraph{Multi-Hop Models.} \citet{chen-durrett-2019-understanding} observe that models without multi-hop reasoning are still able to perform well on a large portion of the test dataset. \citet{hidey-etal-2020-deseption} employ a pointer-based architecture, which re-ranks documents related to a claim and jointly predicts the sequence of evidence sentences and their stance to the claim. \citet{asai2019learning} 
sequentially extract paragraphs from the reasoning path conditioning on the documents extracted on the previous step. CogQA \citep{ding-etal-2019-cognitive}
detect spans and entities of interest and then run a BERT-based Graph Convolutional Network for ranking. \citet{sr-mrs} 
perform semantic retrieval of relevant paragraphs followed by span prediction in the case of QA and 3-way classification for fact checking. \citet{zhou-etal-2019-gear, kgat,trans-xh} model documents as a graph and apply attention networks across the nodes of the graph. We use \citet{trans-xh}'s model due to its strong performance in multi-hop QA on the HotpotQA dataset, in evidence-based fact checking on FEVER, and to evaluate its performance on real-world claim evidence reasoning.

\section{Conclusions}
In this paper, we studied the novel task of multi-hop reasoning for fact checking of real-world political claims, which encompasses both evidence retrieval and claim veracity prediction. 
We presented \politihop, the first political fact checking dataset with annotated evidence sentences.
We compared several models on \politihop\ and found that the multi-hop architecture Transformer-XH slightly outperforms BERT in most of the settings, especially in terms of evidence retrieval, where BERT is easily fooled by named entity overlaps between the claim and evidence sentences. 
The performance of Transformer-XH is further improved when retrieving more than two evidence sentences and the number of hops larger than one, which corroborates the assumption of the multi-hop nature of the task. 
\bibliography{ijcai21}
\bibliographystyle{named}
\newpage
\appendix
\section{Annotation Process}
\label{ref:annotation}

\subsection{Annotation Pipeline}
We used the PolitiFact API to retrieve the articles, along with the source pages used in the article, the claim and the author of the claim. For each article we performed the annotation process as follows:

\begin{enumerate}[nosep]
\item{Reading the claim and the article.}
\item{Picking the evidence sentences from the corresponding PolitiFact article. These sentences should sum up the whole article while providing as much evidence as possible.}
\item{Deciding on the veracity label.}
\item{Going to each relevant url and checking whether it contains the equivalent textual evidence.}
\end{enumerate}

In Step 2, we retrieved evidence sentences sentences, where each sentence follows from the previous one and together they constitute enough evidence to verify the claim and provide an explanation for it.

In Step 4, we wanted to examine how often the evidence can be retrieved from external sources, i.e. not relying on PolitiFact articles. However, we have not gathered enough data to carry out a reliable evaluation of this and thus left the idea for future work.

Originally, following \citep{fever}, we wanted to have a `not enough evidence' label, but due to a small frequency of this label in the annotations, as well as due to a significant disagreement between annotators on that label, we decided to discard it and re-label it with one of the remaining labels (false, half-true or true).
In case of conflicting label annotations, a third annotator was asked to resolve the conflict.

\subsection{Inter-Annotator Agreement}
\begin{figure}
\center
\includegraphics[scale=0.42]{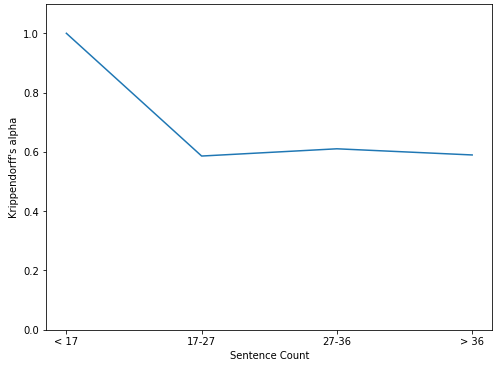} 
\caption{Article length vs. inter-annotator agreement.}
\label{figure:iaa}
\end{figure}

We report Inter-Annotator Agreement (IAA) agreement on the test set, where we had two annotator annotating each instance. For the veracity prediction task, annotators’ Krippendorf's $\alpha$ and Fleiss' $\kappa$ are equal to 0.638 and 0.637 respectively. By comparison, \citet{fever} reported Fleiss' $\kappa$ of 0.684 on the veracity label prediction, which is the another indication of the increased complexity when predicting veracity of claims occurring naturally. For the sentence prediction task, when treating each ruling article as a separate dataset and averaging over all articles, annotators achieve 0.437 Fleiss' $\kappa$ and 0.437 Krippendorff's $\alpha$ (we also compute IAA when treating all sentences from all articles as one dataset, where both IAA measures drop to 0.400). Figure~\ref{figure:iaa} confirms the intuition that annotators tend to agree more on the shorter articles, which are easier to annotate as they contain fewer sets and fewer hops per set. 

\section{Training Details}
We used LIAR-PLUS and \politihop\ for training in three different settings:\\
1. Training on LIAR-PLUS only.\\
2. Training on \politihop\ only.\\
3. Pre-training on LIAR-PLUS and fine-tuning on \politihop. \\
4. Pre-training on FEVER, then fine-tuning on LIAR-PLUS and \politihop.

In the first setting, the models are trained for 4 epochs on LIAR-PLUS. In the second setting, the models are trained for 8 epochs on \politihop. In the third setting, models are trained for 4 epochs on LIAR-PLUS, followed by 4 epochs on \politihop. In every setting, models are evaluated on the dev set and the model with the best label prediction macro-F1 score is saved, which enables early stopping.
For the fourth setting, we pre-train the model for 2 epochs on the FEVER dataset, followed by 4 epochs on LIAR-PLUS, the fine-tune on \politihop\ for 4 epochs.

The models have been trained and evaluated using one NVIDIA TITAN RTX.
We report the results based on a single run with a random seed fixed to 42.

Both BERT and Transformer-XH are trained with the same hyperparameters as in \citep{trans-xh}: BERT 12 layers' with the hidden size of 768, 3 GAT layers with the hidden size of 64. Optimized with Adam with the learning rate of 1e-5.

Some of the experiments are ablation studies of the number of GAT layers and the number of retrieved evidence sentences. The former varies between 1 and 7 while the latter varies between 1 and 10.

\begin{table}[h]
\fontsize{8.4}{8.4}\selectfont
\begin{center}
\begin{tabular}{|p{7cm}|}
\toprule
\textbf{Claim}: Claim: Says 20 million Chinese converted to Islam after it’s proven that the coronavirus doesn’t affect Muslims.\\ 
\textbf{Speaker}: Viral image\\
\textbf{Label}: false \\
\midrule
\textbf{Ruling Comments}: 
[1] Amid fears about the coronavirus disease , a YouTube video offers a novel way to inoculate yourself: convert to Islam. [2] ``20m Chinese gets converted to Islam after it is proven that corona virus did not affect the Muslims,'' reads the title of a video posted online Feb. 18 (...) \textbf{[5] That’s because the footage is from at least as far back as May 26, 2019, when it was posted on Facebook with this caption: 
``Alhamdulillah welcome to our brothers in faith.'' [6] On Nov. 7, 2019, it was posted on YouTube with this title: ``MashaaAllah hundreds converted to Islam in Philippines.''} [7] Both posts appeared online before the current outbreak of the new coronavirus, COVID-19, was first reported in Wuhan, China, on Dec. 31, 2019. \textit{[8] But even if the footage followed the outbreak, Muslims are not immune to COVID-19, as the Facebook post claims. [9] After China, Iran has emerged as the second focal point for the spread of COVID-19, the New York Times reported on Feb. 24 .} [10] 
``The Middle East is in many ways the perfect place to spawn a pandemic, experts say, with the constant circulation of both Muslim pilgrims and itinerant workers who might carry the virus.'' [11] On Feb. 18, Newsweek reported that coronavirus ``poses a serious risk to millions of inmates in China’s Muslim prison camps.'' \\
\bottomrule
\end{tabular}
\end{center}
\caption{An example from the \politihop\ dataset. Each example consists of a claim, a speaker (author of the claim), a veracity label and a PolitiFact article with the annotated evidence sentences. One of the evidence sets is in bold, and the other in italics.}
\label{tab:Example}
\end{table}

\section{Additional Results}

\textbf{Number of evidence sentences vs. evidence retrieval performance}. One of the challenges of generating fact checking explanations is deciding on the length of the explanation. By design, the explanations should be short, ideally just a few sentences. On the other hand, they have to provide a comprehensive motivation of the fact checking verdict. Transformer-XH handles evidence retrieval by ranking the importance of each input sentence. 
We, therefore, pick the most highly ranked sentences, according to the model. By default for all experiments in Section~\ref{sec:results}, the top 6 sentences are used, as it is the average length of an annotation in the \politihop\ test set.

Table~\ref{table:evi-num} shows how recall trades off against precision and improves as an increasing number of sentences is selected. 
Test set F1 grows as the number of sentences grows, while the best train set F1 is the highest for 3 sentences and it gets worse as the number of sentences increases. F1 on dev set is much more even for different numbers, but it peaks at 6 sentences. Generally, 6 sentences gives the best trade-off between the performance on test and dev sets, while being short enough to be considered as short summary of the whole article.



\begin{table}[t]
\footnotesize
\centering
\setlength{\tabcolsep}{4pt}
    \begin{tabular}{lrrr|rrr}
      \toprule
      & \multicolumn{3}{c|}{\textbf{Test}} & \multicolumn{3}{c}{\textbf{Dev}}\\
    \bf  \#S. &\bf F1 &\bf Recall &\bf Precision &\bf F1 & \bf Recall & \bf Precision \\
      \midrule
      1 & 15.9 & 9.7 & \bf 62.0 & 17.3 & 13.5
      & 30.5 \\
      2 & 27.6 & 19.4 & 61.0 & 27.8 & 28.9
      & \bf 31.2 \\
      3 & 36.2 & 28.5 & 61.2 & 31.8 & 39.4
      & 30.0  \\
      4 & 41.3 & 35.5 & 59.1 & 32.3 & 47.0
      & 27.1 \\
      5 & 44.0 & 41.0 & 55.8 & 31.6 & 52.9
      & 24.5 \\
      6 & 47.2 & 47.2 & 54.5 & \bf 32.4 & 60.9
      & 23.8 \\
      7 & 49.2 & 52.5 & 52.9 & 31.8 & 65.8
      & 22.4 \\
      8 & 50.3 & 56.9 & 51.0 & 31.2 & 70.5
      & 21.2 \\
      9 & 50.8 & 60.7 & 49.0 & 30.5 & 74.7
      & 20.2 \\
      10 & \bf 51.0 & \bf 64.1 & 47.3 & 29.1 & \bf 76.4
      & 18.9 \\
      \bottomrule
    \end{tabular}

  \caption{\politihop\ evidence retrieval results for a model trained on LIAR-PLUS $\mathtt{full}$, then fine-tuned on \politihop\ $\mathtt{full}$, with a varying number of top sentences retrieved as evidence. Best number of sentences emboldened.}
  \label{table:evi-num}
\end{table}

\begin{table}
\small
  \begin{center}
    \begin{tabular}{lrrr|rrrrr}
      \toprule
      & \multicolumn{3}{c|}{\textbf{Dev}} & \multicolumn{3}{c}{\textbf{Test}}\\
      & \bf Lab & \bf Evi & \bf Joint &
      \bf Lab & \bf Evi & \bf Joint\\
      \midrule
      BERT & 72.2 & 43.7 & 11.3
      & 72.0 & 43.5 & \bf 14.5\\
      TXH & \bf 73.7 & \bf 44.4 & \bf 12.1
      & \bf 72.6 & \bf 44.6 & 13.4\\
      \bottomrule
    \end{tabular}
  \end{center}
  \caption{LIAR-PLUS results when trained on the LIAR-PLUS full articles dataset. Best model emboldened. The Lab(el) and Evi(dence) results are F1 scores, and Joint is measured with FEVER score.}
  \label{table:liar}
\end{table}

\textbf{LIAR-PLUS}. Here, we investigate training then testing on LIAR-PLUS. As Table~\ref{table:liar} shows, Transformer-XH outperforms BERT by a small margin. This confirms the results in \citep{trans-xh} that Transformer-XH generally performs well in multi-hop settings.

\textbf{Loss function comparison}. In this experiment we compare BERT and Transformer-XH performance on the \politihop\ full article setting when trained with three different loss functions. The default loss function is the sum of evidence prediction loss and label prediction loss. The EVI setting uses evidence loss only and saves the model with the highest validation set evidence f1 score. The LAB setting uses label prediction loss only and saves the model with the highest Macro F1 score on validation data label prediction, just like in the default setting.

Table~\ref{table:loss} shows the best performance with the joint loss for BERT. EVI setting hurts the label prediction while LAB setting hurts the evidence prediction, without providing a clear boost in the second metric over the joint model. Transformer-XH performs much worse on evidence prediction when trained using label prediction loss only. Interestingly, there is no clear performance difference between EVI and default settings.

\begin{table*}[t]
  \begin{center}
    \begin{tabular}{lrr|rr|r|rr|rr|rr|rr|r}
      \toprule
      & \multicolumn{5}{c|}{\textbf{Dev}} & \multicolumn{5}{c}{\textbf{Test}}\\
      & \multicolumn{2}{c|}{\bf Label} & \multicolumn{2}{c|}{\bf Evidence} & \bf Joint &
      \multicolumn{2}{c|}{\bf Label} & \multicolumn{2}{c|}{\bf Evidence} & \bf Joint\\
      & \bf F1 & \bf Acc & \bf F1 & \bf Prec & \bf FEVER &
      \bf F1 & \bf Acc & \bf F1 & \bf Prec & \bf FEVER\\
      \midrule
      BERT & 64.4 & 75.9 & 29.6 & 21.7 & 28.4
      & 57.8 & 79.5 & 45.1 & 52.2 & 23.5\\
      Transformer-XH & \bf 64.6 & \bf 78.7 & 32.4 & 23.8 & \bf 38.3
      & 57.3 & 80.5 & \bf 47.2 & \bf 54.5 & \bf 24.5\\
      BERT-EVI & 34.0 & 51.1 & 31.8 & 23.4 & 26.2
      & 40.7 & 63.5 & 45.7 & 52.8 & 21.5\\
      Trans-XH-EVI & 56.0 & 68.1 & \bf 34.4 & \bf 25.2 & 33.3
      & 59.9 & 75.5 & 46.7 & 54.2 & 21.5\\
      BERT-LAB & 59.4 & 72.3 & 19.9 & 14.8 & 14.9
      & 60.3 & 77.5 & 33.8 & 40.1 & 13.5\\
      Trans-XH-LAB & 62.0 & 75.2 & 18.2 & 13.5 & 14.9
      & \bf 60.6 & \bf 81.5 & 32.4 & 38.8 & 13.0\\
      Random & 24.2 & 27.7 & 14.7 & 12.2 & 0.7
      & 34.1 & 38.5 & 22.9 & 30.2 & 4.5\\
      \bottomrule
    \end{tabular}
  \end{center}
  \caption{\politihop\ results trained on LIAR-PLUS + \politihop\ full articles datasets. EVI means the model was trained with loss on the evidence prediction task only. LAB means the loss on the label prediction only. The default loss was the sum of both. Best model emboldened.}
  \label{table:loss}
\end{table*}

\textbf{Adding sentence IDs to sentence encodings}.
The main goal of this experiment was to see whether providing the information about the positions of the sentences in articles can be leveraged to improve the performance of BERT and Transformer-XH models.

We took the models pre-trained on LIAR-PLUS without sentence positions and fine-tuned the model on \politihop\ with sentence positions, by prepending each sentence's encoding with the token [unusedN], where $N=$sentence position.
Table~\ref{table:sent-ids} shows a significant performance boost for Transformer-XH label prediction, but not for evidence retrieval. BERT does not exhibit any improvement, which is to be expected as it considers each sentence in isolation, it doesn't learn any interactions between sentences.\\

\begin{table*}[t]
  \begin{center}
    \begin{tabular}{lrr|rr|r|rr|rr|rr|rr|r}
      \toprule
      & \multicolumn{5}{c|}{\textbf{Dev}} & \multicolumn{5}{c}{\textbf{Test}}\\
      & \multicolumn{2}{c|}{\bf Label} & \multicolumn{2}{c|}{\bf Evidence} & \bf Joint &
      \multicolumn{2}{c|}{\bf Label} & \multicolumn{2}{c|}{\bf Evidence} & \bf Joint\\
      & \bf F1 & \bf Acc & \bf F1 & \bf Prec & \bf FEVER &
      \bf F1 & \bf Acc & \bf F1 & \bf Prec & \bf FEVER\\
      \midrule
      BERT & 62.4 & 75.2 & 27.7 & 20.3 & 24.1
      & 57.4 & 79.0 & 42.9 & 49.2 & 21.5\\
      Transformer-XH & \bf 65.2 & \bf 78.0 & \bf 32.2 & \bf 23.6 & \bf 38.3
      & \bf 66.5 & \bf 84.0 & \bf 47.2 & \bf 54.9 & \bf 26.5\\
      \bottomrule
    \end{tabular}
  \end{center}
  \caption{\politihop\ results for training on LIAR-PLUS $\mathtt{full}$, then fine-tuning on \politihop\ $\mathtt{full}$ with sentence ID encodings. Best model emboldened.}
  \label{table:sent-ids}
\end{table*}

\textbf{Label confidence vs. performance}.
The goal here was to measure how confident the models are in their label predictions and to see if higher confidence means better performance. The results are presented in Table~\ref{table:lab-conf}.

Transformer-XH is usually sure of its predictions, so not much can be observed based on that - it does indeed have higher F1 score when it is more confident, but there are too few instances where it is not confident to make any conclusions - apart from the one that it's often sure but makes a mistake anyway. Besides, NE overlap was not particularly high for the instances where the model got confused.

The effect is even stronger with BERT to the point where having less than 95\% confidence usually results in a bad prediction.

\begin{table*}[t]
  \begin{center}
    \begin{tabular}{lrr|rr|r|rr|rr|rr|rr|r}
      \toprule
      & \multicolumn{5}{c|}{\textbf{Transformer-XH}} & \multicolumn{5}{c}{\textbf{BERT}}\\
      & \multicolumn{2}{c|}{\bf Label} & \multicolumn{2}{c|}{\bf Evidence} & \bf Joint &
      \multicolumn{2}{c|}{\bf Label} & \multicolumn{2}{c|}{\bf Evidence} & \bf Joint\\
      &\bf  F1 &\bf Acc &\bf F1 &\bf Prec &\bf FEVER &\bf
      F1 & \bf Acc &\bf F1 &\bf Prec &\bf FEVER\\
      \midrule
      $<$ 90\% & \bf 45.5 & \bf 46.2 & \bf 51.6 & \bf 39.7 & \bf 34.6
      & 30.9 & 32.6 & 52.9 & 41.6 & 20.9\\
      $\geq$ 90\% & 67.2 & 78.3 & \bf 54.1 & \bf 40.7 & 67.0
      & \bf 72.5 & \bf 85.7 & 50.9 & 37.8 & \bf 67.3\\
      \bottomrule
    \end{tabular}
  \end{center}
  \caption{\politihop\ adversarial dev set performance vs. label confidence.}
  \label{table:lab-conf}
\end{table*}

\end{document}